# Navigational Path-Planning For All-Terrain Autonomous Agricultural Robot

Vedant Ghodke

*Abstract*— The shortage of workforce and increasing cost of maintenance has forced many farm industrialists to shift towards automated and mechanized approach. The key component for autonomous systems is the path planning techniques used. Coverage path planning (CPP) algorithm is used for navigating over farmlands to perform various agricultural operations such as seeding, ploughing, or spraying pesticides and fertilizers. This report paper compares novel algorithms for autonomous navigation of farmlands. For reduction of navigational constraints, a high-resolution grid map representation is taken into consideration specific to Indian environments. The free space is covered by distinguishing the grid cells as covered, unexplored, partially explored and presence of obstacle. The performance of the compared algorithms is evaluated with metrics such as time efficiency, space efficiency, accuracy, and robustness to changes in the environment. Robotic Operating System (ROS), Dassault Systemes Experience Platform (3DS Experience), MATLAB along with Python were used for the simulation of the compared algorithms. The results proved the applicability of the algorithms for autonomous field navigation and feasibility with robotic path planning.

## I. INTRODUCTION

Path Planning is an important primitive for autonomous mobile robots that lets robots find the shortest – or otherwise optima l – path between two points, here two points on the field or fa rm. Path planning is a navigational problem to find a path of valid configurations to move the robot from the source to destination. Path-planning requires a map of the environment and the robot to be aware of its location with respect to the map and robot types such as cleaner robots [3], lawn mower robots [1], mining robots [4], painting robots [2] and many more applications.

The main goals of the motion and path planning of this project are: 1. Introduction of suitable map representations 2. Implementation of path-planning algorithms ranging from Dijkstra to A*, D*, RRT, Potential Fields, Fuzzy Logic 3. Variations of the path-planning problem for better efficiency and accuracy.

The 2-dimensional mapping of the robot can be discussed in brief with the help of the summarized points below:

1. Robot should have complete and non-repetitive coverage
2. Navigation should include all possible points with ROI
3. Avoidance of path overlapping
4. Obstacle avoidance and path alternating
5. Simple trajectory generation for coverage and navigation
6. Addressing of all navigational, time and space constraints to find the most optimal path

Over the past decades, several researchers have developed navigational algorithms for various agricultural applications. Navigational algorithms focused for agricultural applications have been focused on specific target regions of a particular farmland.

The global agriculture output of India, as of 2018, stands at 7.68 percent. After People's Republic of China, India is the second largest producer in this field. There has been a massive increment in the demand and contrastingly, the application of technology into this sector is much lower as compared to several other nations. This has served as the inspiration for the development of independent field navigational algorithms and thus, robots for the purpose agricultural path planning.

The entire review paper hereafter is organized in the following manner. Section B includes a literature review on the existing methodologies. Section C briefs about the farm representation technique proposed for achieving autonomous navigation. In Section D, the algorithm comparison for autonomous navigation is presented. Section E discusses the actual validation of the compared algorithms. Section 6 includes the conclusion, results and future scope of the research and implementational work.

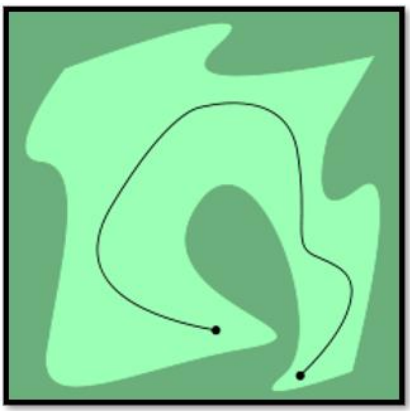
**Valid Path**

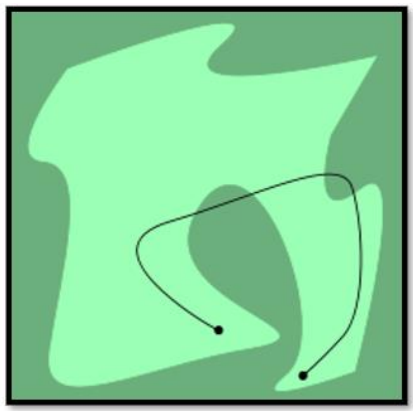
**Invalid Path**

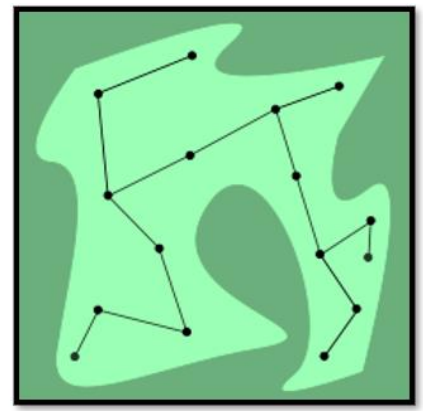
**Possible Path Outcomes**

## II. RELATED WORKS

Navigation algorithms' separation from path planning techniques was proposed in [7]. They were classified as online and offline systems based on the methodology used for achieving complete coverage. The robotic vehicle is proposed to be completely autonomous and not dependent on any infrastructure such as radio beacons or GPS sensor. It can operate below the vegetation canopy, which poses a problem for drone-based solutions. While operating in farm, robotic vehicle will collect data related to crops, builds map and routes in real time. To navigate in the field, it uses 3-D vision camera, LiDAR and compass sensors along with an on-board AI computing system. The vision camera will be mounted on tilt and span assembly, which will provide 360-degree image capture capability. The path planning algorithms have been studied, and most suitable algorithm will be used for implementation based on its convergence to finding optimal path to reach the goal. The robot will learn the environment in which it is moving and navigates between the points on the map it creates [8] and use this approach. The strategy used in this paper is to find the most optimal and yet the most cost-effective path planning solution for navigation of agricultural fields.

Few of the methods proposed in literature review for performing navigation include a standard template model method [9], genetic algorithms [10], back-turn method [11], neural network method [12], probabilistic algorithms, bidirectional RRTs and internal spiral coverage [13].

In the template model map method, a set of pre-decided map templates are used to achieve navigation, and a mismatch between the template map and the real-world surroundings leads to poor accuracy and coverage of the algorithm. This is not true for all agricultural operations since field structure is not constant as it changes with topography and climatic changes in Indian environments. Several techniques rely on heuristics and randomization for navigation[14]. They are efficient in terms of equipment used, but generate several paths that overlap and have high time complexity to achieve navigational coverage.

The rate of navigational coverage is an increasing function of the length of the path to be traversed. Covering more areas require more time and complexity dealing. Navigational paths overlap and the average traveling velocity for coverage are the vital factors for determining the performance of the coverage algorithm.

Path terrain is directly proportional to the mean velocity of the robot, since sharp corners and hurdles tend to slow down the robot performing coverage. Many algorithms proposed for navigational process assume that robots have access to its precise orientation and position. This is satisfied by the different SLAM algorithms (simultaneous localization and mapping) [15].

Compared to the common systems, SLAM techniques require a very high cost system for mapping and visualization. In this paper too, this factor has been taken into consideration, since the objective is to compare the most optimum algorithms for the process of navigational traversal.

## III. FARM REPRESENTATION

Navigational algorithms achieve fulfillment by navigating the robot into an uncovered farm space by using several techniques of decomposition in the free space into position cells and perform navigational coverage for each individual. The techniques utilized include approximation, semi-approximation and exact decomposition parameters. In semi-approximate navigation, the farm search space is partially fragmented into positional cells, with position cells having a fixed width with variable top and bottom. Exact decomposition navigation generates a set of non-intersecting cells, the combination of which would fill the target navigational space. Each cell would be covered by motions in back-and-forth manner.

Semi-approximation navigational techniques break-down the environment into positional grid cells of roughly uniform size. A typical positional grid cell is represented by a square. Each grid cell is assigned a specific value to represent the obstacles and free navigational space for traversal. Grid based methods are complete for navigation and a re hence preferred. The resolution of the grid map affects the total navigational coverage completed.

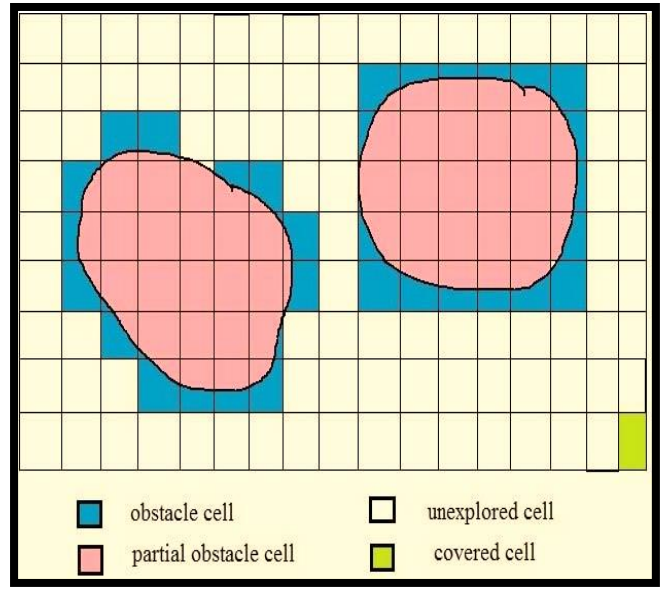

Fig. 1. Segregation of positional grid cells

The agricultural and farm land in India, varies in size and shape. A standardized set of template map models may not be suitable for all Indian environments. In this paper, a grid based positional representation is proposed for the farm navigation coverage algorithms.

Since the map resolved image affects the computational time and accuracy the in path planning as well, several algorithms restrict from using high resolution maps of the environment. This is because, autonomous robotic equipment used in agriculture is not capable of making real- time adjustments in the decision-making process. A binary value is associated with each positional grid cell to help in planning the navigational coverage path. Figure 1 (above) shows an example of a grid based farm mapping of an environment with obstacles and terrain changes. A self developed online platform is considered for planning coverage simulation. The simulated robot is given the task of identification of boundaries and obstacles.

## IV. NAVIGATIONAL PATH PLANNING

Navigational coverage algorithms many-a-times implement on the assumption that the farm-to-grid cell is equal in size as that of the robot. Thus, if a robot visits a positional cell then it ensures that it completely covers it. This factor makes many navigational algorithms computationally non-expensive. However, the drawback is that the areas around the obstacles are not covered. As the free space around the obstacle is also within the grid, they are ignored as obstacle which would make the coverage ratio high is achieved by the compared algorithms.

However, in real time there would be areas not covered near obstacles. This would increase as the number of obstacles in the agricultural environment increases. The compared algorithms overcome these drawbacks by making the positional grid size larger than the size of the actual robot. For the purpose of mapping, the farm position cells are classified as unexplored, obstacle, partial obstacle, and covered. At initial stage, each cell would be marked as unexplored. If a position cell is within the robot's boundary, then it is marked as covered. Cells with obstacles are marked as obstacle category.

Mapping of the farm fields is done as a background process while conducting the navigational and mapping motion. Figure 2 represents the flow and transition chart of the robot's operation. In the beginning, a robot can either be new to the field or may begin coverage after an interpretation of the field. If the robot is in a completely new farm region, then localization and mapping of the robot is performed. Once the region has been fully mapped, the map is converted into a perceivable format. There are several scenarios where the system restarts due to various factors such as traversal issues, commutation power, hindrances from the field terrain etc.

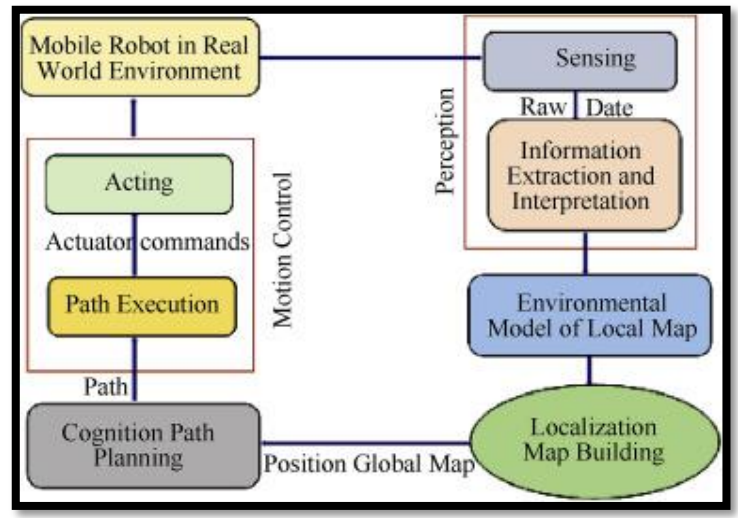

**Fig. 2. State transition diagram for the robot's operation**

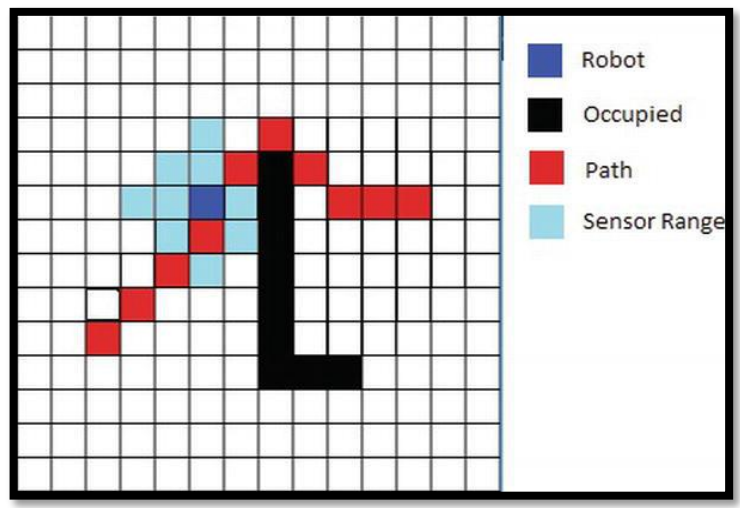

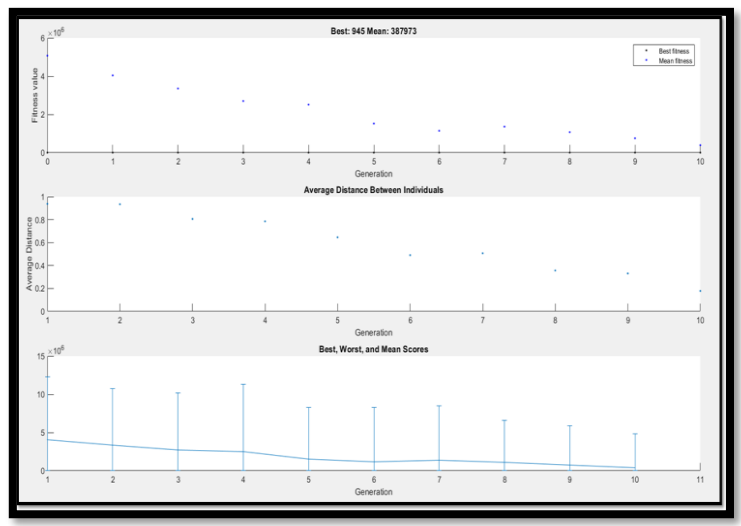

**Graphical representation of Genetic Algorithm on the map layout of the field**

The overall procedure of the navigational algorithm is depicted in Fig. 3. At an initial stage, all the cells are marked as unexplored. After the probable estimation, navigational path generation is done. If an obstacle is recognized in the field during this phase then choosing of different suitable algorithm is initiated. When path generation is complete, distance calculation function is used to select the next optimum position, which is at minimum distance from the current robot position. The robot moves to the selected position and the simultaneous path mapping procedure continues. The above procedure takes in real-time for the robot to be able make real-time decisions.

The distance calculation function operates in real-time by evaluating the non-occupied positions around the robot.

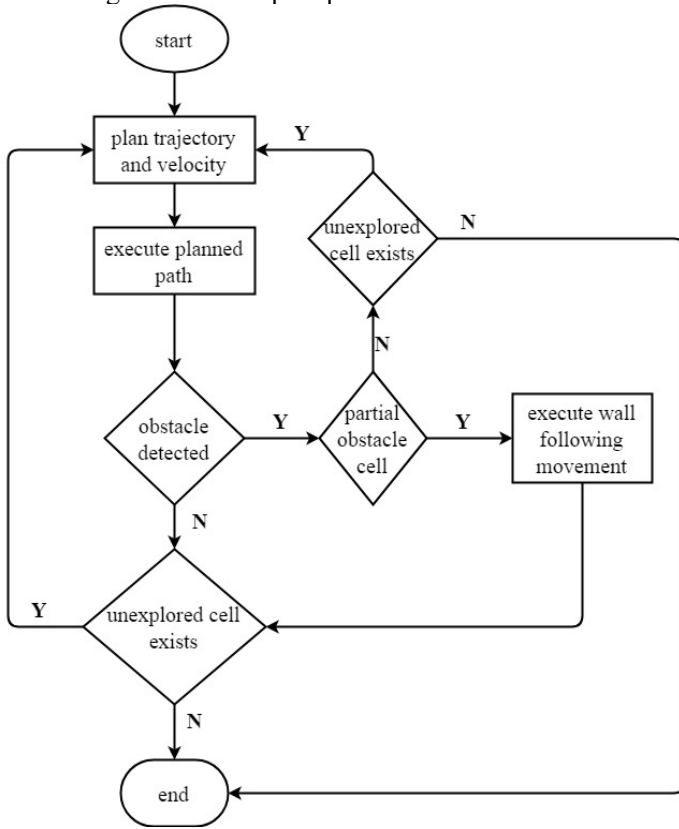

**Fig. 3. Flow of the Coverage path planning procedure**

### V. SIMULATIONAL VALIDATION

Simulated navigation mappings were conducted on a system with 8 GB of RAM, 4 GB graphic memory, Windows 10 operating system. Robotic Operating System (ROS) kinetic and Dassault 3DS Experience platform applications paired with MATLAB R2020a and Python were used to simulate the initial robot movements. A custom robot chassis was created in the 3DS platform as well to help perform several tasks along with path-planning simultaneously. The environment modeling was done with obstacles to test the actual performance with sample maps and obstacles. A map with three obstacles inside the environment was considered for evaluation.

| Map with two irregular obstacles | Without partial cell | | With partial cell | | |
|---|---|---|---|---|---|
| | Time taken (in minutes) | Coverage (%) | Number of partial cells identified | Time taken (minutes) | Coverage (%) |
| Run 1 | 7.40 | 77 | 15 | 17.30 | 89.52 |
| Run 2 | 10.22 | 83 | 13 | 16.03 | 88.05 |
| Run 3 | 8.57 | 79 | 17 | 19.51 | 91.71 |

**Table 1: Comparison of time taken and coverage achieved for the given map with and without partial obstacle cell categorization in MATLAB R2020a**

The designed robot is equipped with high-grade equipment which aids initial mapping and localization of the robot with other tasks too. Table 1 above represents the navigational coverage percentage achieved. Two different scenarios were considered for evaluation.

*Scenario 1:* Navigational coverage simulation was performed by segregating the farm as unexplored, covered and obstacle presence.

*Scenario 2:* Farm positions were classified as unexplored, covered, obstacle, and partial obstacle presence.

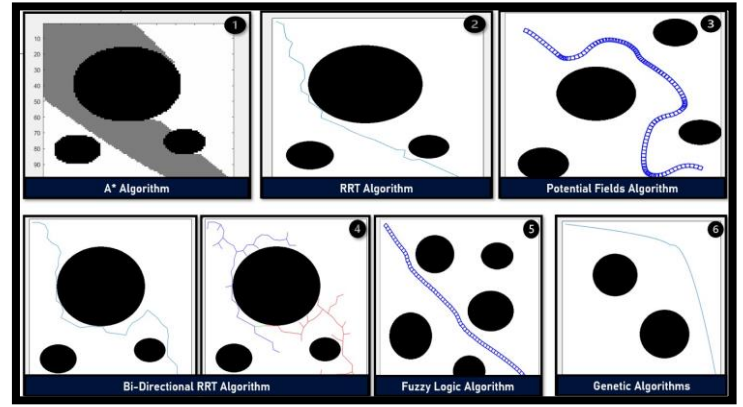

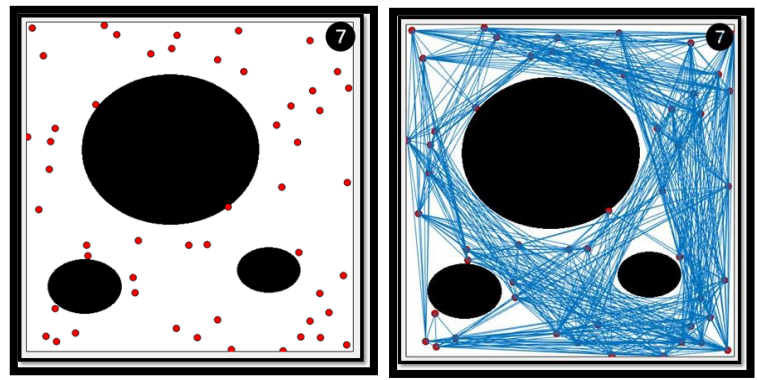

**Probabilistic Path Planning Implementation in MATLAB R2020a**

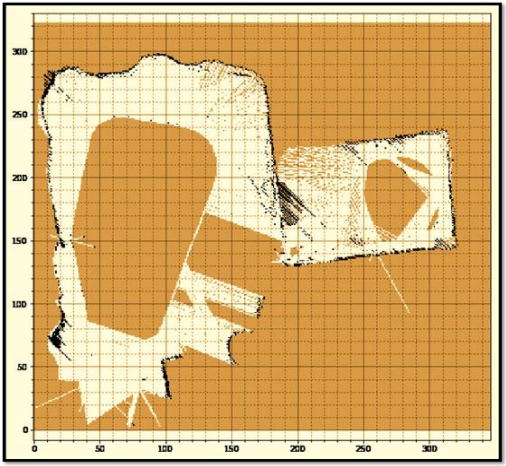

**Fig. 4. Occupancy Grip Map for navigation of the robot**

Scenario 1 took less time to achieve complete farm navigation but the total distance traversal achieved was less since, the areas around obstacles were not taken into consideration. This was due to the methodology of classification of farm grid cells used. Scenario 2 has an extra classification of cells as partial obstacle cell. This allows in coverage of those areas as well. Figure 4 shows heuristic a visualization of the navigational map performed by the robot. As noted in table 1, the time taken to achieve total coverage has increased.

However, the main objective is to achieve total navigational coverage of thew field. This is the chief requirement of autonomous navigation in agriculture. Since this was implemented in ROS and 3DS platforms coupled with MATLAB R2020a, it can be directly interfaced with an agricultural machinery in real time.

## VI. CONCLUSION

A detailed comparison of the different path-planning algorithms for navigating agricultural fields as been conducted. Farm position grid cells were classified as covered, unexplored, presence of obstacles and partial obstacles. Possible paths were initiated to achieve navigational coverage in unexplored positions and in path-planning motion to achieve coverage around obstacles as well. Distance calculation function was used to calculate the shortest and optimal distance from current position to the adjacent unexplored positional cells and those with least cost was selected as the robot's next location for navigation.

Though the duration to achieve optimum coverage was increased due to switching between different path- planning algorithms, the overall coverage percentage was increased due to the visiting of the partially occupied farm grid cells. This is the decisive performance implementation factor for agricultural path planning. The path- planning evaluation was done by representing the 3-dimensionalenvironment in 2-dimensional space which reflect the practicality of algorithms for real time robotic navigation in agricultural fields.

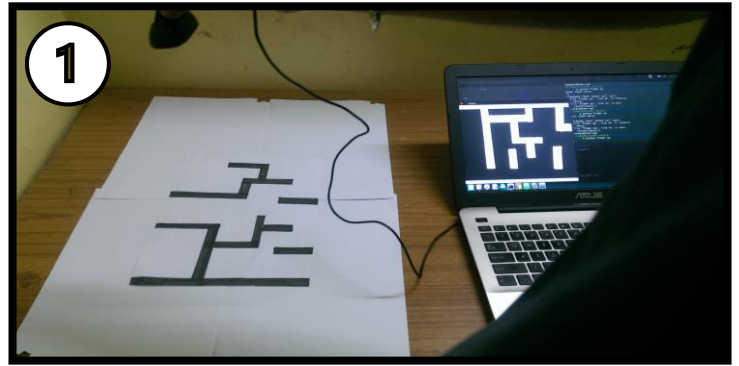

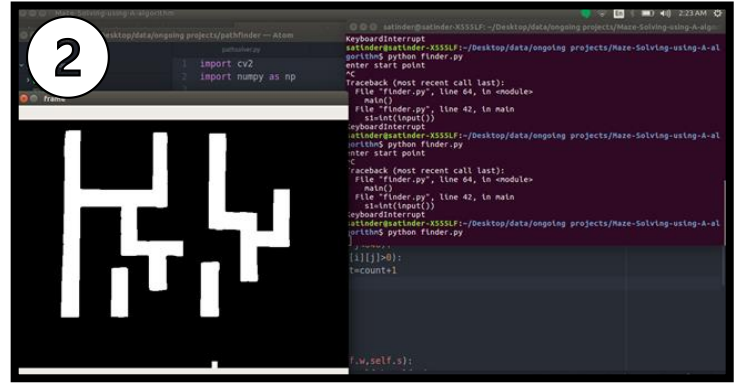

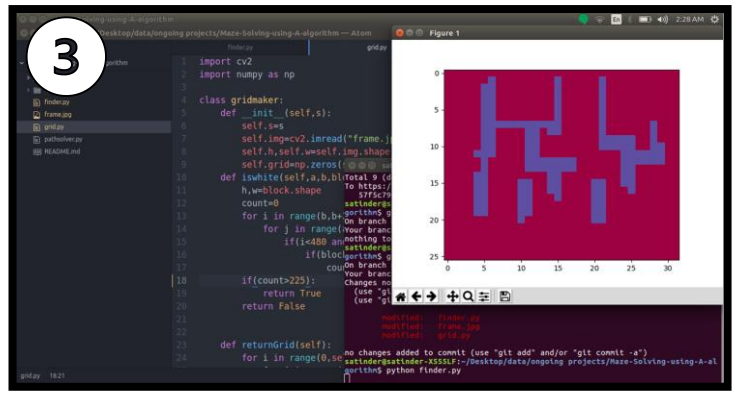

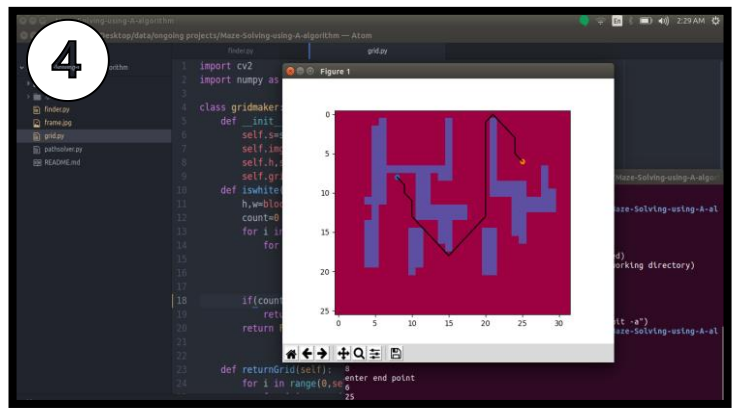

**Real-Time Implementational Testing in Python environment**

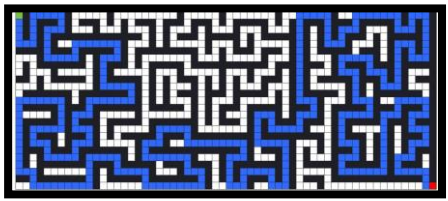

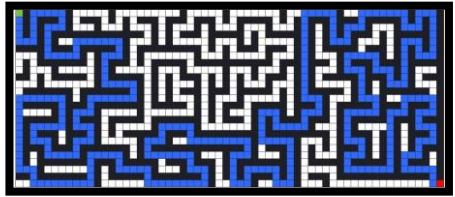

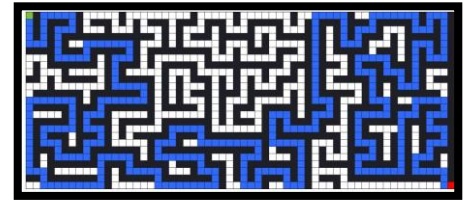

**A* Algorithm**  **Dijikstra's Algorithm**  **DFS Algorithm**

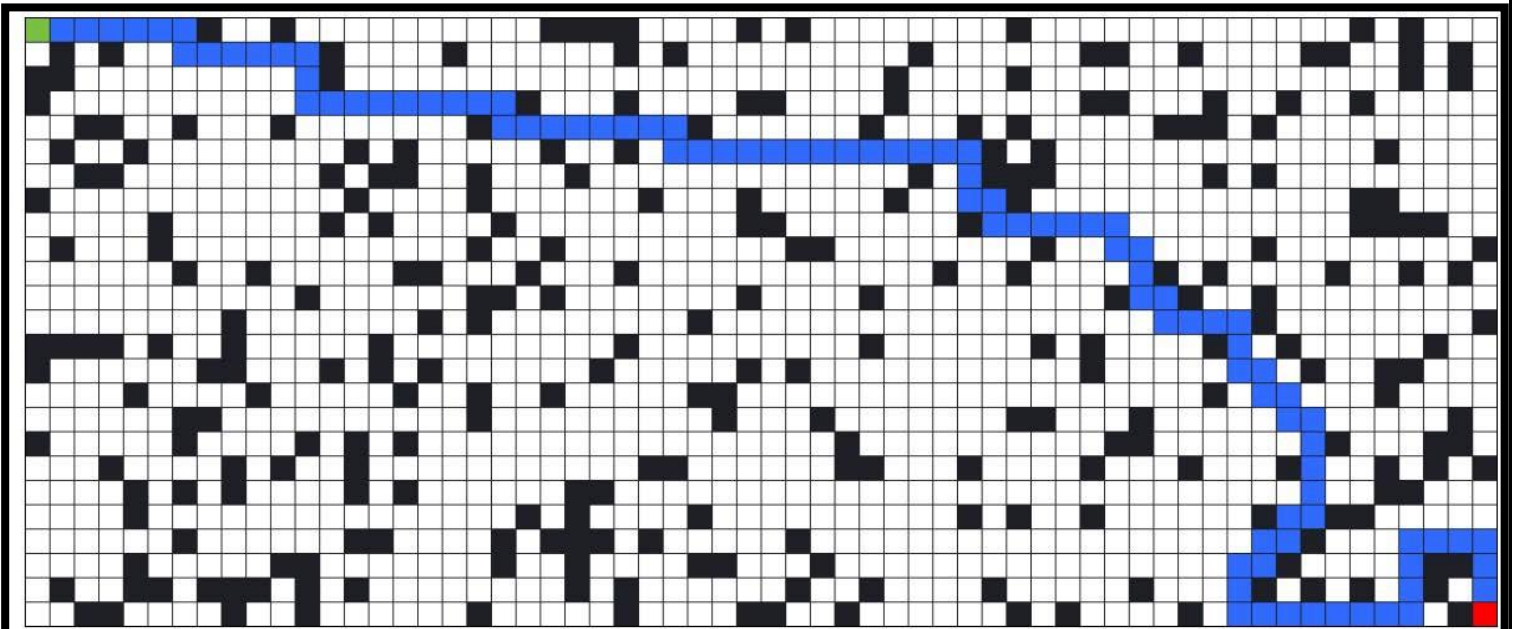

**A* Algorithm - Random Farm/Field**

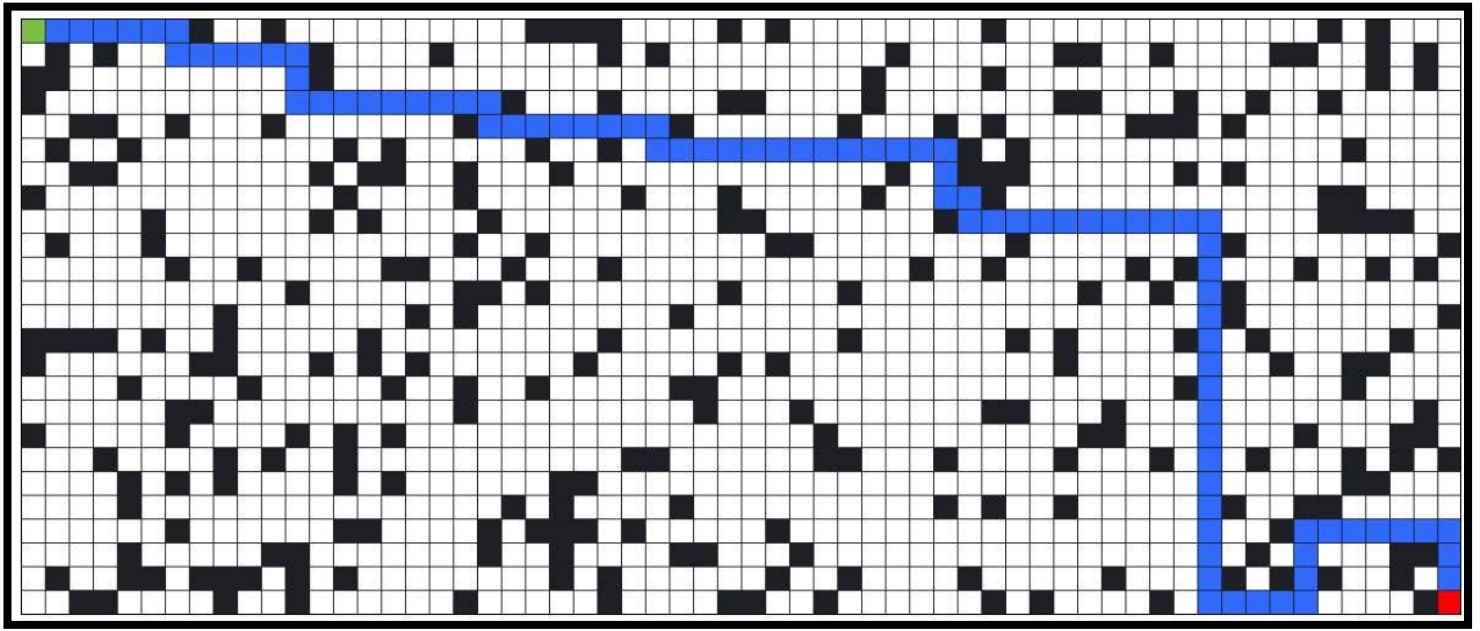

**Dijikstra's Algorithm - Random Farm/Field**

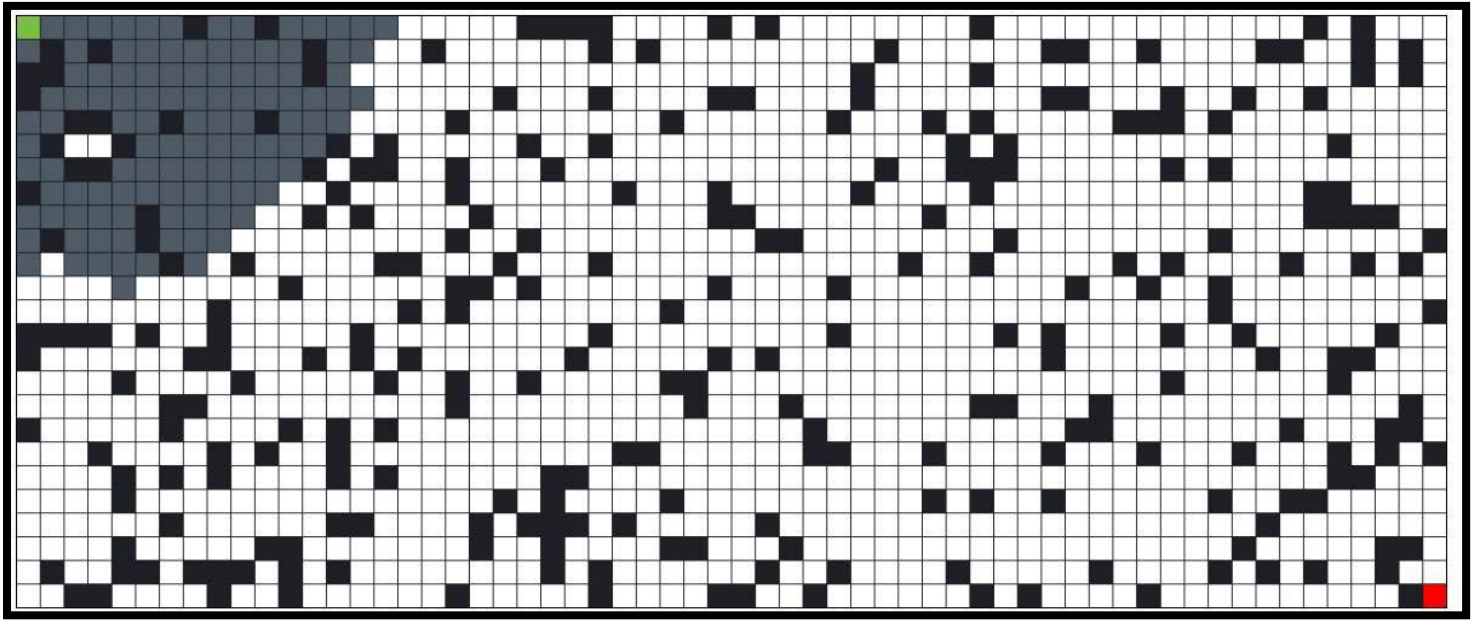

**Depth-First-Search Algorithm - Random Farm/Field**